# Automated LoD-2 Model Reconstruction from Very-High-Resolution Satellite-derived Digital Surface Model and Orthophoto

*Shengxi Gui [a,b], Rongjun Qin [a,b,c,d,*]*

[a] *Geospatial Data Analytics Lab, The Ohio State University, Columbus, USA;*

[b] *Department of Civil, Environmental and Geodetic Engineering, The Ohio State University, Columbus, USA;*

[c] *Department of Electrical and Computer Engineering, The Ohio State University, Columbus, USA;*

[d] *Translational Data Analytics Institute, The Ohio State University, Columbus, USA;*

*\*Corresponding author:* qin.324@osu.edu, *2036 Neil Avenue, Columbus Ohio 43210, USA.*

**Abstract:**

Digital surface models (DSM) generated from multi-stereo satellite images are getting higher in quality owing to the improved data resolution and photogrammetric reconstruction algorithms. Very-high-resolution (VHR, with sub-meter level resolution) satellite images effectively act as a unique data source for 3D building modeling, because it provides a much wider data coverage with lower cost than the traditionally used LiDAR and airborne photogrammetry data. Although 3D building modeling from point clouds has been intensively investigated, most of the methods are still ad-hoc to specific types of buildings and require high-quality and high-resolution data sources as input. Therefore, when applied to satellite-based point cloud or DSMs, these developed approaches are not readily applicable and more adaptive and robust methods are needed. As a result, most of the existing work on building modeling from satellite DSM achieves LoD-1 generation. In this paper, we propose a model-driven method that reconstructs LoD-2 building models following a "decomposition-optimization-fitting" paradigm. The proposed method starts building detection results through a deep learning-based detector and vectorizes individual segments into polygons using a "three-step" polygon extraction method, followed by a novel grid-based decomposition method that decomposes the complex and irregularly shaped building polygons to tightly combined elementary building rectangles ready to fit elementary building models. We have optionally introduced OpenStreetMap (OSM) and Graph-Cut (GC) labeling to further refine the orientation of 2D building rectangle. The 3D modeling step takes building-specific parameters such as hip lines, as well as non-rigid and regularized transformations to optimize the flexibility for using a minimal set of elementary models. Finally, roof type of building models s refined and adjacent building models in one building segment are merged into the complex polygonal model. Our proposed method has addressed a few technical caveats over existing methods, resulting in practically high-quality results, based on our evaluation and comparative study on a diverse set of experimental datasets of cities with different urban patterns. (codes /binaries may be available under this GitHub page: https://github.com/GDAOSU/LOD2BuildingModel)

**1. Introduction**

Remotely sensed satellite imagery effectively acts as one of the preferred ways to reconstruct wide-area and low-cost 3D building models used in urban and regional scale studies (Brown et al., 2018; Facciolo et al., 2017; Leotta et al., 2019). However, because satellite imagery has a relatively lower spatial resolution than aerial imagery and LiDAR, 3D building models base on satellite data faces challenges in detecting buildings, extracting building boundaries, and reconstructing an accurate building 3D model, especially in regions with small and dense buildings (Sirmacek & Unsalan, 2011). Level-of-Detail (LoD) building models defined through the city geography markup language (CityGML), present 3D building models in several levels 0 to 4 (Gröger et al., 2008). An improved LoD standard CityGML 2.0 speciate LoD-0 to LoD-3 as four sub-definitions from LoD x.0 to LoD x.3 bases on the exterior geometry of buildings (Gröger et al., 2012; Biljecki et al., 2016). With relatively fine digital surface models derived from satellite photogrammetry, using satellite-based DSM and building polygon to generate Level-of-Detail 1 (LoD-1) 3D building models with a flat roof is now becoming a standard practice. However, 3D building models reconstruction with prototypical roof structures (LoD-2), remains a challenging problem, especially from low-cost data sources such as satellite images (Kadhim & Mourshed, 2018; Bittner & Korner, 2018). Despite the increasingly higher resolution of satellite images (as high as 0.3-0.5 GSD) on par with past aerial images, the generated 3D information from the satellite images, due to 1) image quality (being high altitude collection distorted by the atmosphere), 2) relatively low resolution in comparison to aerial data, and 3) cross-track collections (Qin, 2019) with temporal variations, often posses high uncertainties leading to challenges in LoD-2 modeling. Although 3D building modeling from point clouds has been intensively studied, most of the methods are still ad-hoc to specific types of buildings and assume these input point clouds to be highly accurate and dense (i.e. those captured with LiDAR). However, when presented with low-resolution point clouds with relatively high uncertainties as produced by satellite images, these methods no longer produce reasonable results, especially for areas with dense buildings, as traditional bottom-up approaches are not able to identify and separate between individual buildings given the low resolution and blurry boundaries both in height (from digital surface models (DSM) or point clouds) and in color (or spectrum), therefore it requires approaches to be sufficiently robust to identify, regularize and extract individual buildings, often following a top-down approach (i.e. model-driven). To extract individual and well-delineated boundaries of the building, one often needs to decompose "overestimated" building footprint candidates, using regularity, spectral or height cues (Brédif et al., 2013; Partovi et al., 2019), and adapt models that are well-suited to the extracted/decomposed footprints. However, among the limited studies focusing on LoD-2 reconstruction based on satellite-derived DSMs (Alidoost et al., 2019; Partovi et al., 2019; Woo & Park, 2011), caveats exist in the decomposition procedure, which often does not fully explore the use spectrum, DSM and contextual information, resulting in unsatisfactory results, especially in areas where buildings are dense: on the one hand the individual buildings are often incorrectly extracted/oriented, on the other hand, buildings in clusters are not consistent.

In this paper, we revisit this process and fill a number of gaps by developing a three-step approach that specifically aims to improve the extraction of building polygon and fitting models in such challenging cases, which formed our proposed LoD-2.0 model reconstruction pipeline for satellite-derived DSM and Orthophotos: our proposed method starts with building mask detection by using a weighted U-Net and RCNN (region-based Convolutional Neural Networks), followed by the proposed three-step approach through "extraction-decomposition- refinement" for regularized 2D building rectangle generation. In the "extraction" step, we vectorize building masks as boundary lines, and then regularize lines orientation through line segments from orthophoto. In the "decomposition" step, a grid-based building rectangle generating method is developed by a grid pyramid to generate rectangles and subsequent separating and merging steps to optimize polygons. In the "refinement" step, we post-refine the 2D model parameters through propagating similarities of neighboring buildings (in terms of their orientation and type) through a Graph-Cut (GC) algorithm, with optionally a second refinement step using OpenStreetMap road vector data. The individual building rectangle and the corresponding DSMs are then fitted by taking models from a pre-defined model library that minimizes the difference between the model and the DSM. In this process, the motivation is to fully utilize the spectral and height information when performing the decomposition process and take global assumptions on orientation and building type consistencies of the building clusters to yield results that are superior to state-of-the-art methods. The proposed method is validated using a diverse set of data, and although our approach follows and refines the existing modeling paradigm, we find the proposed approach yields robust performance on various types of data and is able to reconstruct areas with dense urban buildings. This therefore leads to our contributions in this paper: 1) we present a model-driven workflow that performs LoD-2 model reconstruction that yields highly accurate results as compared to state-of-the-art methods; 2) We demonstrate that the use of a combined semantic segmentation and Region-based CNN (RCNN) leverages detection and completeness for object-level mask generation; 3) we propose a novel three-stage ("extraction-decomposition-refinement") approach to perform vectorization of building masks that yields superior performance; 4) We validate that the use of multiple cues of neighboring buildings, and optionally road vector maps, can generally improve the accuracy of the resulting reconstruction.

The remainder of the paper is organized as follows: Section 2 introduces related works, and Section 3 describes our approach for 3D building modeling in detail, includes pre-processing, building detection and segmentation, building footprint extraction, building polygon decomposition, building polygon orientation refinement, 3D building model fitting, adjacent building model merging. Section 4 summarizes four study areas in two cities and experiment results of qualitative and quantitative evaluation, and the comparison with other current methods. Finally, Section 5 concludes this paper.

## 2. Related work

There are many approaches for the detection and reconstruction of 3D building models from airborne photogrammetry and LiDAR data (Cheng et al., 2011; Wang, 2013), and relatively few on the satellite-derived data (Arefi & Reinartz, 2013; Woo & Park, 2011) given that dense matching method yielding relatively high-quality data was only getting advanced in recent years (Bosch et al., 2017; Leotta et al., 2019; Qin, 2016; Qin et al., 2019), and SAR data contributes to LOD-1 building reconstruction in state and national scale recent years (Geis et al., 2019; Li et al., 2020; Frantz et al., 2021). Most satellite-based methods can be classified into three categories: data-driven, model-driven, and hybrid approaches. In data-driven approaches, buildings are assumed to be individual parts of roof planes, by considering the geometrical relationship of point, lines, and surface from DSMs and point cloud. In model-driven approaches, buildings DSMs or point clouds are compared with 3D building models in the model library, to select the most appropriate fitting model and the best parameters. In hybrid approaches, both former approaches have been included that data-driven approaches extract the geometric feature (line, plane) of the building model, and model-driven approaches compute the model parameters and reconstruct the 3D model.

Typical LoD-2 model reconstruction approaches using satellite images take the following steps: The initial step detects and segments building areas from either images or combined images and orthophotos (Qin & Fang, 2014). Traditional approaches for building detection used Support Vector Machine (SVM, Gualtieri & Cromp, 1999) to classify building classes as to explore sparse labels, and spatial features exploiting region-wise information such as Length-Width Extraction Algorithm (LWEA, Shackelford & Davis, 2003) are stacked into the feature vectors for classification. Recently, deep learning models are often used to perform so-called semantic segmentation to fully explore the growingly available labeled satellite datasets, among which U-Net (Ronneberger et al., 2015) is often the baseline to start with, which predicts labels for every pixel. On the other hand, instance-level prediction is of relevance, as detectors capable of detecting individual buildings may move a step further for building modeling. The fast region-based Convolutional Network (Fast R-CNN, Girshick, 2015) belongs to this class of approach, which took a window-based approach to efficiently identify regions (bounding boxes) containing objects of interest, and its advanced version, the Mask R-CNN (He et al., 2017) concurrently identifies the regions (bounding boxes) of the individual object of interest, as well as performing pixel-level labeling within each bounding box, which normally serves as a baseline approach for any type of detection tasks, which a few variants of this type of approaches are available as well (Cai & Vasconcelos, 2018; Zhang et al., 2020);

Once initial building masks are extracted, the next logical step is to vectorize the masks such that the boundaries of individual buildings are modeled with regularized polylines (hereafter we call it building boundaries). Initial steps to pre-processing the detected masks can be to use shape reconstruction methods such as alpha-shape (Kada & Wichmann, 2012) or gift wrapping algorithm (Lee et al., 2011) to obtain initial boundaries, which might be followed by line fitting methods to further simplify the lines of the boundaries by using for example, random sampling consensus (RANSAC, Schnabel et al., 2007), least-square line fitting (O'Leary et al., 2005), and Douglas–Peucker algorithm (Douglas & Peucker, 1973). The processes can be aided by extracting lines directly from orthophotos, such as LSD (Line Segment Detector, Von Gioi et al., 2010) and KIPPI (KInetic Polygonal Partitioning of Images, Bauchet & Lafarge, 2018) algorithms. The proposed method adopts a pipeline based on the Douglas-Peucker algorithm to initially extract building polygons from the mask and uses LSD to refine boundary lines. Another refinement method directly refines the orthogonal boundary lines from Mask R-CNN polygon (Zhao et al., 2018). With refined building boundaries, regularizations or decompositions can be performed to further identify individual buildings for fitting preliminary building models: for example, Partovi et al. (2019) proposed an orthogonal line-based 2D rectangle extraction method that assumed orthogonality and parallelism of the building footprint, which aims to decompose a tentative building footprint to rectangle shapes by starting from the longest boundary lines. Once these individual rectangle shapes are extracted, a merging operation might be needed to obtain the final simplified shapes (Brédif et al. 2013). Often this process can be aided with supplementary information such as OpenStreetMap or Ordnance Survey data (Haklay, 2010) when those with sufficient quality are available. Furthermore, Girindran et al. (2020) proposed a 3D model generation approach that uses open-source data directly, including OSM and Advanced Land Observing Satellite World 3D digital surface model (AW3D DSM).

The model fitting for satellite-derived data is usually carried out by fitting a few preliminary models in a model library, based on the resulting 2D building rectangle and the available DSM, and these models normally consist of a few parameters to allow efficient fitting. Partovi et al. (2015) calculated the roof components parameters using an exhaustive search to fit the DSMs. The proposed approach follows a process of a defined model library and model parameters exhaustive searching. Other than exhaustive search, Alidoost et al. (2019) presented a deep learning-based approach to predict nDSM and roof parameters of roof based on multi-scale convolutional–deconvolutional network (MSCDN) from the aerial RGB images. Given the low-resolution (and potentially high uncertainty of the 3D geometry), a post-processing step can be considered to recover complex roofs such as merging of falsely separated building components through for example, eave (Brédif et al., 2013) or ridgeline (Alidoost et al., 2019). This step is often ad-hoc with different levels of regularization depending on the accuracy of the detected 2D building footprint, the complexity of the model and accuracy of the 3D geometry (i.e. DSM). In recent years, end-to-end methods create a solution to derive building models from the remotely sensed image directly. Qian et al. (2021) provided a deep learning method Roof-GAN to generates structured geometry of roof structures.

## 3. Methodology

Fig. 1 presents the workflow of our method. We follow the typical reconstruction paradigm as introduced in Section 2, with a produced DSM and Orthophoto (pre-processing), the process consists of six main steps: 1) building segmentation, 2) building polygon extraction, 3) building rectangle decomposition, 4) orientation refinement, 5) 3D model fitting, and 6) model post-refinement. The input data of our method is DSM and the corresponding orthophoto, which is generated by using a workflow based on RPC stereo processor (RSP) that includes sequentially a level 2 rectification, geo-referencing, point cloud generation, DSM resampling, and orthophoto rectification (Qin, 2016). The building segmentation from orthophoto is developed using a combined U-Net based semantic labeling and Mask R-CNN segmentation algorithm. To extract building polygon, we propose a novel method that first detects the rough boundary of building segments using the Douglas–Peucker algorithm and then adjusts and regularizes the boundary lines combining orientation from LSD. Next, to generate and decompose the basic 2D building model from building polygon, especially in the polygon with several individual small buildings, a novel grid-based decomposition method is developed to generate and post-process the building rectangles with serval sub-rectangles in a building segment. The building rectangles are subject to an orientation refinement process using GC optimization and optionally a rule-based adjustment using OpenStreetMap (OSM). LOD-2 model reconstruction is performed by fitting the most appropriate roof model in a building model library by optimizing the shape parameters that yield minimum RMSE with the original DSM. In the last step, model roof type is refined using GC optimization, and simple building models are merged as complex models based on a few heuristics. The detailed components of each step are given in the following sections.

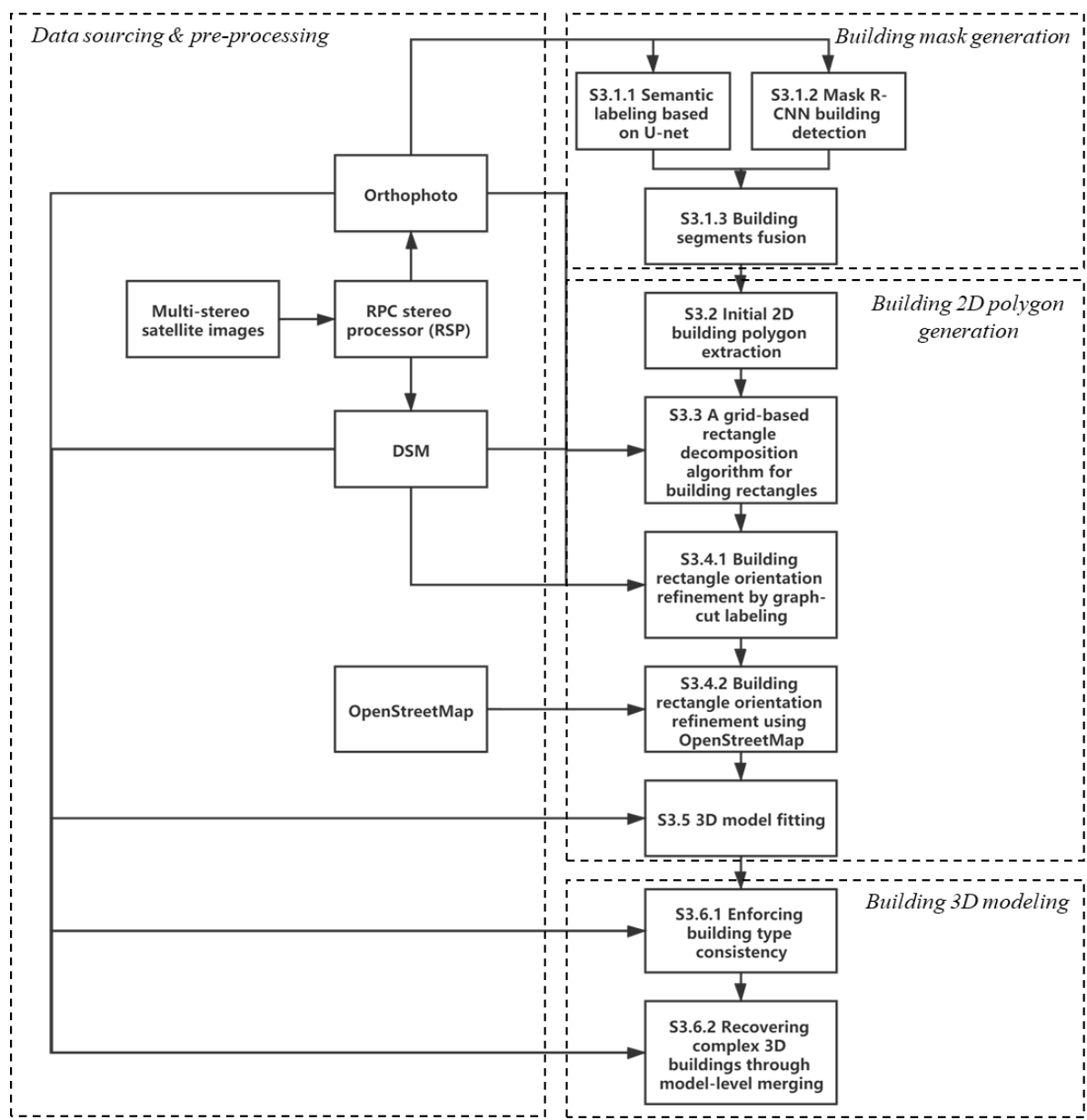

**Fig. 1.** Workflow of our proposed method, each key component is described in subsections in the texts.

*3.1. Building detection and segmentation*

We used an approach that combines two baselines as introduced in Section 2: the U-Net and Mask R-CNN. U-Net (Ronneberger et al., 2015) with its structure designated for preserving details, provides well-delineated boundaries of objects, while Mask R-CNN (He et al., 2017) with its original structure, although less complex to preserve good boundaries, has the ability to perform instance-level segmentation thus often to have better recall, while given the nature of the region-based detection for Mask R-CNN, it may omit certain large-sized buildings. These two are complementary to each other, therefore we perform a segmentation-level fusion for results generated by both networks (introduced in Section 3.1.3). In the following sections, we introduce training details of U-Net, in which we used a revised loss (Qin et al., 2019), and the Mask R-CNN, both of which are trained using satellite datasets.

*3.1.1 Semantic labeling based on U-net*

We used the training dataset provided by John's Hopkins University Applied Physics Lab's (JHUAPL) through the 2019 IEEE GRSS Data Fusion Contest (Le Saux et al., 2019). In total, five classes are considered in this data: ground, tree, roof, water, and bridge and elevated road. While in practice, those five objects are not balanced, so the number of training patches is adjusted to make the number of each category have similar patches (Qin et al., 2019).

For the data preparation of training, to improve the training samples, each image with a size of 1024 × 1024 pixels was split into 512 × 512 patches with 50% overlap. Thus, for each image of 1024 × 1024, 9 patches are available for semantic segmentation so that the training samples are nine times larger than the original samples.

The loss for this network is defined as dynamic and class-weighted, to adjusts weights of five classes based on their number of pixels during the process in the individual patch. The weight is computed as:

$$W_i = N_c \cdot (1 - \frac{n_i}{N}) \quad (1)$$

where $W_i$ means the loss weight for class $i$, $N_c$ represents the number of class, $n_i$ is the number of pixels in the current patch, and $N$ is the total valid pixels in the current patch.

For the prediction part, the data splitting and fusion method are performed as well. The original testing images with a size of 1024 × 1024 are divided into 512 × 512 patches with 80% overlap. Moreover, four predictions are performed by rotating the patches into four directions for each patch. The final prediction can be derived by fusing these predictions with the following strategy: the size of each patch is 512*512, a buffer with a width of 64 pixels in the patch border will not be considered (weight = 0), otherwise pixels will have a weight equal to 1. In the merging process, a voting strategy is used to calculate the summed weight of each patch (predictions of four directions are rotated to the original direction), and the class of pixels will be assigned as the class with the highest weight in that pixel. Thus, the segmentation result can be developed from orthophotos.

*3.1.2 Mask R-CNN building detection*

The only category we need is building, and this group is masked for training, validating, and testing. The vertexes of building masks in samples are extracted using the building polygon extraction method, described in Section 3.2 and Section 3.3. Furthermore, the bounding-boxes are generated from these vertexes as well.

In the study, the pre-train weight focus on building segmentation provided by CrowdAI (Mohanty, 2018) is used to Mask R-CNN processing. Furthermore, the dataset of 2019 IEEE GRSS Data Fusion Contest (Le Saux et al., 2019) is used for training and validation, to achieve a specific weight of network for building segments of orthophoto. During the training step, a total number of 120 epochs training has been performed, with 30 epochs for RPN training, 60 epochs for FCN training, and 30 epochs for all layers. Thus, each epoch contains 500 training steps and 50 validation steps. Fig. 2 shows the training process for Mask R-CNN for a different component of the loss. There is a small disturbance in the epoch of 30 (between RPN training and FCN training), and it remains stably decreasing afterward. On the other hand, in the validation part, only RPN loss slightly increase, and the other losses include Mask R-CNN loss still remain decreasing with respect to the training loss.

For the prediction part, the data splitting and fusion method are performed as well. The original testing images with a size of 1024 × 1024 are divided into 512 × 512 patches. Moreover, the entire segmentation is developed by merging the segment of the individual patch.

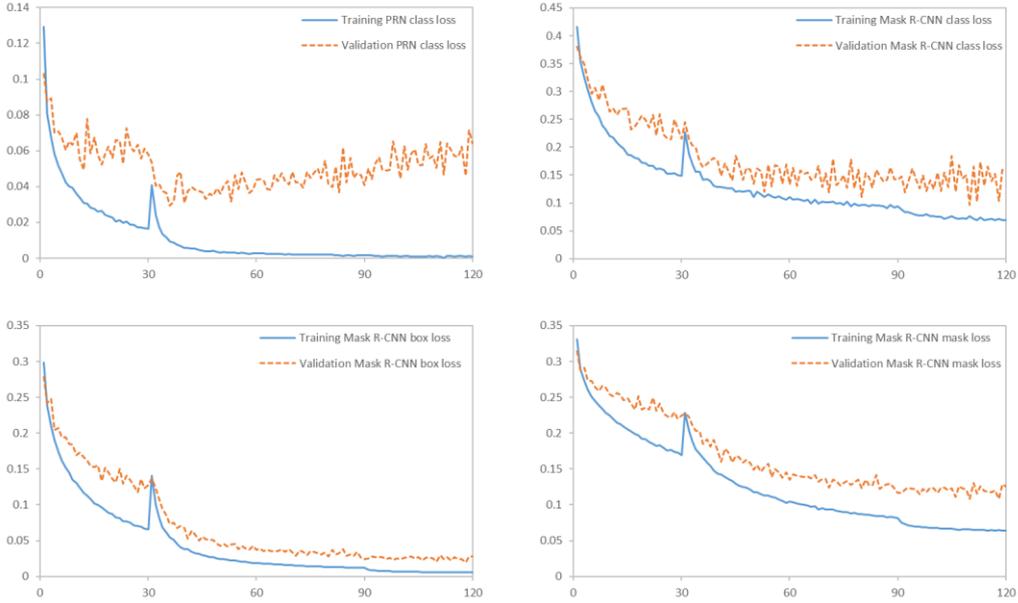

**Fig. 2.** The relationship between epochs and Mask R-CNN loss for training and validation

*3.1.3 Building segmentation fusion*

As mentioned above (a schematic workflow is shown in Fig. 3), we used a segmentation-level fusion to obtain the initial building masks. The fusion method is rather heuristic, as U-Net tends to produce sharp object boundaries and Mask R-CNN tends to capture individual objects. Therefore, we start by using the U-Net result as the primary segmentation and use a decision weight $w$ to decide if the label of a segment from Mask R-CNN needs to be used for replacing the result of UNet, as shown below:

$$w = \frac{1}{area_{bbox}} \cdot \frac{area_{class}}{area_{bbox}} = \frac{area_{class}}{(area_{bbox})^2} \qquad (2)$$

where $w$ indicates the decision weight of a bounding box, $area_{class}$ is the area of the masked pixel cover in the bounding box, $area_{bbox}$ is the area of the bounding box. The rationale of this formulation follows the observation that the Mask R-CNN tends to perform well with smaller buildings (possibility due to a large number of small buildings in the training set), and as long as detection has a good filling within its bounding box, it may be used as a confident detection of individual buildings. The determination threshold is empirical (here we variably used 0.1 to 0.2 to yield reasonable results for this 0.5 m resolution data), which is stable in one region. If the average size is large, the threshold will be close to 0.1. Otherwise, the threshold will be close to a larger value. , shows the decision process. Firstly, use the semantic labeling result as primary classification. Secondly, calculate the segment weight of each segment from Mask R-CNN result, select segments with weight exceeding threshold as potential segments. Finally, use the potential segments in Mask R-CNN result to replace the corresponding mask of semantic labeling result. 3, shows the decision process. Firstly, use the semantic labeling result as primary classification. Secondly, calculate the segment weight of each segment from Mask R-CNN result, select segments with weight exceeding threshold as potential segments. Finally, use the potential segments in Mask R-CNN result to replace the corresponding mask of semantic labeling result.

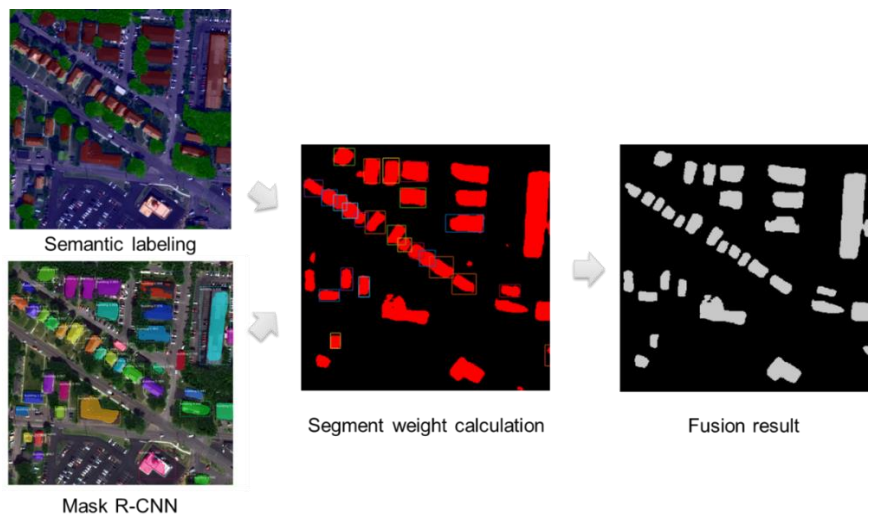

**Fig. 3.** A schematic workflow for segmentation-level fusion

*3.2 Initial 2D building polygon extraction*

Building polygon extraction aims to vectorize building boundaries as polylines to assist regularized building model fitting. These polylines need to follow constraints as man-made architectures do, such as complying with orthogonality and parallelism. For coarse resolution data, this is simplified as polylines consisting of parallel or orthogonal lines, the extraction of which in this work follows three steps: initial line extraction, line adjustment, and regularization; the result of each step by example is shown in Fig. 4.

The ***initial line extraction*** is performed with well-known Douglas–Peucker algorithm (Douglas & Peucker, 1973). It starts at the boundary points of the building mask and extracts lines recursively. Due to the irregularity of the building masks, the initial line segments present errors, sharp angles and are short. The ***line adjustment*** process aims to use the concept of main orientations to connect and turn these short line segments into consistently long and straight lines: One or more main orientations are defined for the polygon to separate fitted lines into bins presenting orthogonal lines pairs. First, we draw a histogram of orientations varying from 0° to 90° with 10° of the interval, and each bin presents two angles that are 90° different: for example, for the bin present 30°, it would include line segments close to -60° as well. This is to facilitate maximizing the line adjustment to contain mainly orthogonal and parallel lines for regularity. Then we sum up the length of lines for each bin and select the bins whose summed length is bigger than a given threshold (here we set 120 pixels), in which we conclude the main orientation by taking the weighted average (weight being the length of each segment) of the orientations in that bin. The orientation for the bin is computed as the weighted average of lines' orientation in this group (Note: orthogonal orientations are deemed identify thus only represented by angles within 0°-90°). The adjustment is made by iterating through each line segment, reassigning to the closest main orientation, intersect and reconnect with neighboring segments; clear differences can be noted in Fig. 4(a) and Fig. 4(b).

The ***line regularization*** process further optimizes line orientation by utilizing detected lines from the orthophoto: Line-Segment Detector (LSD) algorithm (Von Gioi et al., 2010) is used to extract line segments and the orientation of each line segment is used to readjust orientations of the nearest line of the initially extracted line segments from the building mask to enable consistencies between the detected footprint and the image edges. The resulting building footprint after this process is regarded as the initial building polygon for reference.

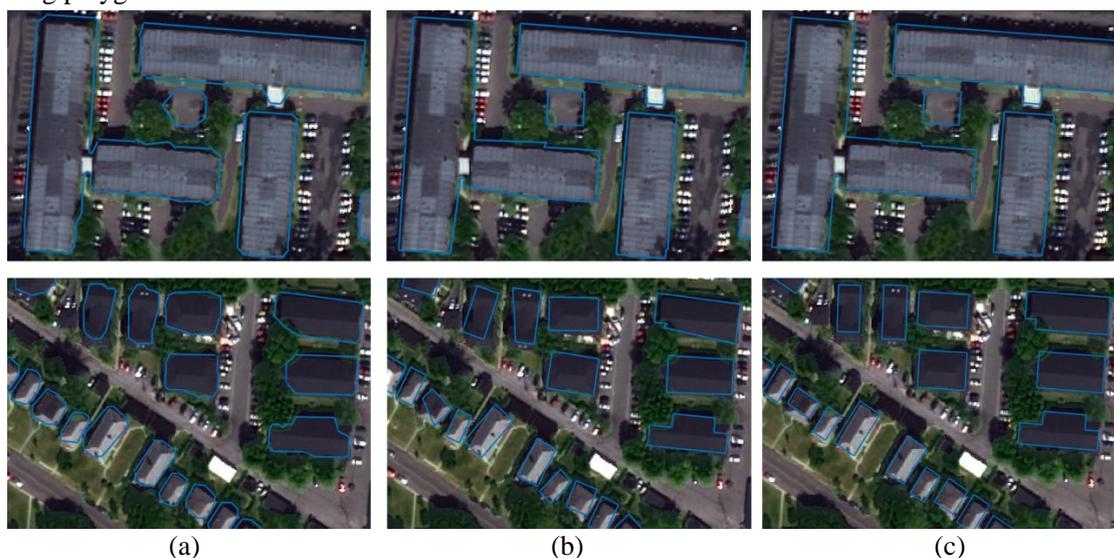

           (a)                              (b)                            (c)

**Fig. 4.** Two examples are showing boundary extraction of building polygon. (a) boundary after initial line extraction; (b) boundary after line adjustment; (c) boundary after line regularization.

*3.3. A grid-based rectangle decomposition algorithm for building rectangles*

    The polyline based initial building polygon can be overcomplex for the limited number of simplex models to fit, thus it requires to decompose these building polygons (as produced following method introduced in Section 3.2), into simple shapes, which in this work as basic rectangles on which the model fitting can base on (we call it building rectangle hereafter). Existing methods mostly follow a greedy strategy for decomposition, which identifies the main orientation following the longest straight line, extends parallel lines gradually to get decomposed rectangles to fill the polygon (Kada, 2007; Partovi et al., 2019). However, often the results of such a greedy-decomposition method do not match the actual building components. We therefore present our grid-based decomposition approach, in which can be flexibly integrated information from the orthophoto and DSMs to guide this process.

    The grid-based decomposition approach assumes that multicomplex building polygons generally follow the Manhattan type and are composed of simplex rectangles, the seamlines of which can be approximately either horizontal or vertical on a 2D grid. The workflow is shown in Fig. 5. To start, we first rotate the 2D building polygon by aligning its main orientation to the x-axis (Fig. 5), and the same transformation will be performed on the Orthophoto and DSM. The decomposition of the polygon to rectangles starts with a DSM and orthophoto based initial decomposition: first, gradients of the DSM in both the horizontal and vertical directions are computed and the those larger than a threshold (0.3 m), followed by a non-maximum suppression (with a window size of 7 pixels) are selected as the candidate of lines for separating the polygon, which is further filtered by considering the color information: for each candidate separating line, we create a small buffer on both side and only lines with color differences of between these two buffers bigger than a threshold $T_d$ (we used 10 for an 8-bit image). For the segments after this initial decomposition, a maximum inner rectangle extraction (Alt et al., 1995) is performed to extract individual rectangles. Note this above process is performed on the coarsest layer of an image pyramid to reduce impacts of the noise, and the results are then interpolated to the finer layers for further analysis. We kept three layers for our 0.5m resolution data, (grid level 1-3, Fig. 5), meaning that the top

layer is ¼ of the original resolution.

The individual rectangles will then be projected to the original resolution grid and post-processed by using a merging operation, as the above process might over separate the building masks. However, here the merging operation will only be performed on rectangles that share one edge, as the merged rectangles must be rectangles as well to enable fitting. To start, we first identify adjacent rectangles by dilating their boundaries for 7 pixels and define adjacency as long as there are overlaps and the common edge has a similar length (the difference shorter than 5 pixels in length). The criterion used to decide if two adjacent rectangles should be merged as below:

$$\begin{cases} merge, \ |\overline{C_1} - \overline{C_2}| < T_d \ \cap \ |\overline{H_1} - \overline{H_2}| < T_{h1} \cap max|\Delta H_{edge}| < T_{h2} \\ not \ merge, \ otherwise \end{cases} \quad (3)$$

meaning that the mean color differences ($|\overline{C_1} - \overline{C_2}|$) of the two rectangles (projected onto the orthophoto) is smaller than a threshold $T_d$ (here we define as 10 for an 8-bit image); 2) mean height difference ($|\overline{H_1} - \overline{H_2}|$) is smaller than a threshold $T_{h1}$ (1 m); 3) There is no dramatic height changes in a buffered region that cover the common edge, to avoid narrow streets in between, and this criterion is defined as the height gradient ($\Delta H_{edge}$) in the overlapped region (after the dilation) smaller than $T_{h2}$ (0.2 m).

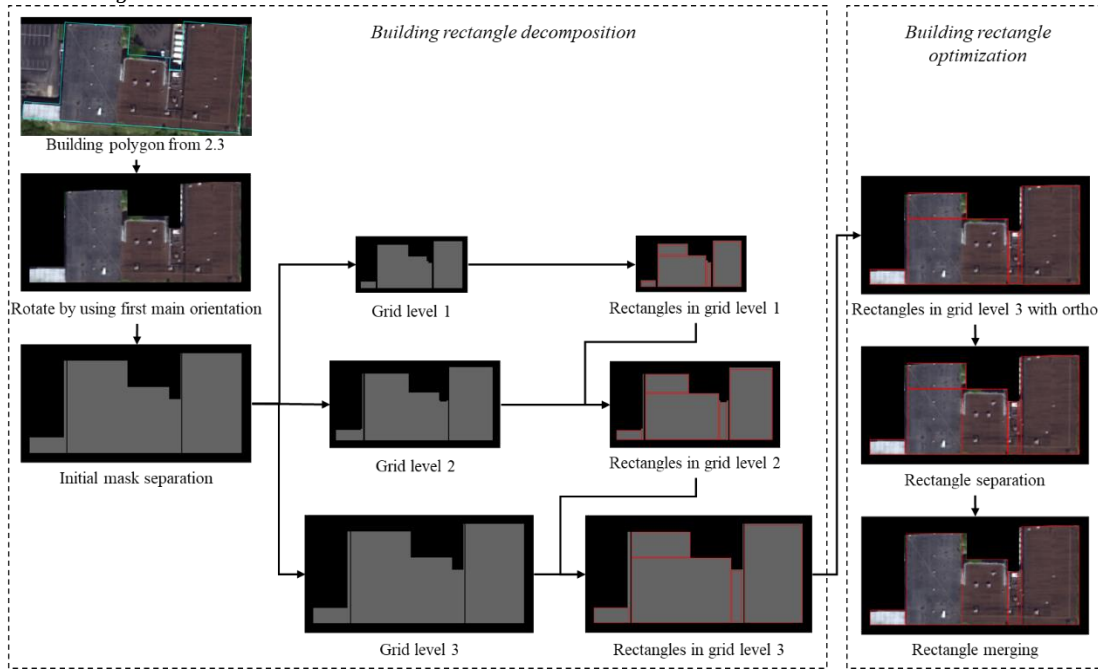

**Fig. 5.** The proposed grid-based decomposition algorithm. Details are in the texts.

*3.4 Orientation refinement for Building rectangles*

The rectangular building elements produced based on the process as described above are independent of each other, and on the other hand, the orientation of these rectangle footprints can be easily impacted by noises of the initial building mask and the orthophotos & DSM, considering that most neighboring buildings follow consistent orientations, such as those in the same street block. Therefore, we optimize the orientations for buildings using the Graph-Cut algorithm (Boykov & Jolly, 2001), and as an optional step to refine the orientation using available OpenStreetMap road networks (covering 80% of the cities worldwide (Barrington-Leigh & Millard-Ball, 2017)).

*3.4.1 Building rectangle orientation refinement by graph-cut labeling*

We formulate the orientation adjustment problem as a multi-labeling problem, by assigning each building rectangle, possible labels as discrete angle values, ranging from 0° to 180°, with a 2° as the interval, resulting in approximately 90 labels (0° pointing to the north). We aim to assign these labels such that similar neighboring building rectangles have the same or similar labels, where similarity is defined by using texture and height differences. GC is well known for solving multi-label problems in polynomial time (Kolmogorov & Zabih, 2002). It aims to minimize a cost function consisting of data and a smoothness term, in which the smoothness terms aims to enforce consistency between nodes and the data terms encode a priori information, shown in

$$E(\mathcal{L}) = \sum_{x_i \in P} R(x_i, \mathcal{L}_i) + \lambda \sum_{(x_i, x_j) \in \mathbb{N}} B(x_i, x_j) \delta(\mathcal{L}_i, \mathcal{L}_j) \qquad (4)$$

$$R(x_i, \mathcal{L}_i) = 1 - e^{-|\theta_{x_i} - \theta_{\mathcal{L}_i}|} \qquad (5)$$

$$B(x_i, x_j) = W(i,j) \qquad (6)$$

$$\delta(\mathcal{L}_i, \mathcal{L}_j) = \begin{cases} 0, if\ \mathcal{L}_i = \mathcal{L}_j \\ 1, if\ \mathcal{L}_i \neq \mathcal{L}_j \end{cases} \qquad (7)$$

where $\mathcal{L} = \{0,1,2,\ldots,89\}$ is the label space indicating the orientation of the rectangle being $\theta_\mathcal{L} = 2 \times \mathcal{L}$ degrees, and $\mathcal{L}_i$ refers to the optimized label (or $\theta_{\mathcal{L}_i}$ as the optimized orientation) for the building rectangle $x_i$. $\theta_{x_i}$ refers to the initial orientation of the building rectangle and the data term $R(x_i, \mathcal{L}_i)$ tends to keep the initial orientation unchanged and is defined as a normalized value (smaller than 1) that gets bigger as the optimized orientation differs more. $B(x_i, x_j)$ is the smooth term, which defines the similarity of neighboring building rectangles (set $\mathbb{N}$), and $\lambda$ is the weight that leverages the contributions of the smoothness term. The smoothness term $B(x_i, x_j)$ is determined by using affinity matrix $W(i,j)$, and which adapts an exponential kernel (Liu & Zhang, 2004):

$$W(i,j) = e^{-dist(i,j)/2\sigma^2} \qquad (8)$$

$$dist(i,j) = \sqrt{\begin{aligned}w_r(\boldsymbol{nr_i} - \boldsymbol{nr_j})^2 + w_\theta(n\theta_i - n\theta_j)^2 + w_S(\boldsymbol{nS_i} - \boldsymbol{nS_j})^2 \\ + w_C(\boldsymbol{nC_i} - \boldsymbol{nC_j})^2 + w_\sigma(\boldsymbol{n\sigma_i} - \boldsymbol{n\sigma_j})^2\end{aligned}} \qquad (9)$$

where $W(i,j)$ is the affinity weight matrix for neighboring building rectangle $i$ and $j$, with $0 < W(i,j) < 1$, $\sigma$ is the bandwidth of exponential kernel. $dist(i,j)$ is the similarity between building rectangle $x_i$ and $x_j$, calculated as a weighted (and normalized) combination of a few factors including the distance of two rectangle center $\boldsymbol{nr_i}$, the difference of the orientation, the shape $\boldsymbol{nS_i}$, the color $\boldsymbol{nS_i}$, and that of the color variations $\boldsymbol{n\sigma_i}$, respectively defined as $\boldsymbol{nr_i} = (nX_i, nY_i)$, where $nX_i$ and $nY_i$ referring to the location of the rectangle center, $\boldsymbol{nS_i} = (nL_{1,i}, nL_{2,i})$ referring to the length $nL_{1,i}$ and width $nL_{2,i}$ of the rectangle, $\boldsymbol{nC_i} = (nC_{R,i}, nC_{G,i}, nC_{B,i})$ referring the mean value of RGB that rectangle covers, $\boldsymbol{n\sigma_i} =$ $(n\sigma_{R,i}, n\sigma_{G,i}, n\sigma_{B,i})$ referring to the standard deviation of RGB. We give a higher weight value $w_r = 3$ to location, to ensure consistencies are mainly optimized locally; the weight of orientation $w_\theta$ and the weight of shape $w_S$ are set by default as 1, and the weight of RGB color $w_C$ and the weight of RGB standard deviation $w_\sigma$ are set as 0.3, since both the $nC$ and $n\sigma$ have three components. Note all measures for computing similarities are normalized through an exponential kernel similar to (8). These weights (for location, orientation & shape, color-consistency to decide similarity of two neighboring buildings), are set to be intuitive and often can be fixed in the optimization based on the quality of the data. It is possible to automatically tune these parameters based on prior information of the dataset with the meta-learning method, or to decide the parameter based on the distribution of the data (while only slight difference from our empirical values (Kramer, 1987)). Here since these parameters are intuitive and explainable, these weights do not proceed an automated tuning.

*3.4.2 Building rectangle orientation refinement using OpenStreetMap*

OpenStreetMap is a publicly available vector database and it was known to have covered more than 80% (Barrington-Leigh & Millard-Ball, 2017) of the road networks globally, thus it may serve as a valid source to refine building orientations for the well geo-referenced dataset. We made a simple assumption that the buildings always have the same direction as their surrounding road vectors (Zhuo et al., 2018). Therefore, we perform the OSM based orientation refinement following simple heuristics or reassigning orientation using nearest criterion: for each building rectangle, the nearest road vector is found by computing the distance between the center of the building rectangle and each road vector, and if the intersection angle between the main orientation line of a building polygon (introduced in Section 3.2) and the nearest road vector is smaller than 30°, the orientation of the rectangle will be adjusted, otherwise kept unchanged. This process will allow minor adjustment of the orientation when the OSM data is available. Two examples are shown in Fig. 6 as comparisons when the GC optimization and the OSM-based adjustment are respectively applied, resulting in adjusted rectangles (in blue and green). Once OSM is unavailable in a certain region or orientation difference between building and OSM street line, orientation refinement using OSM will not execute and there will be sole refinement by using GC.

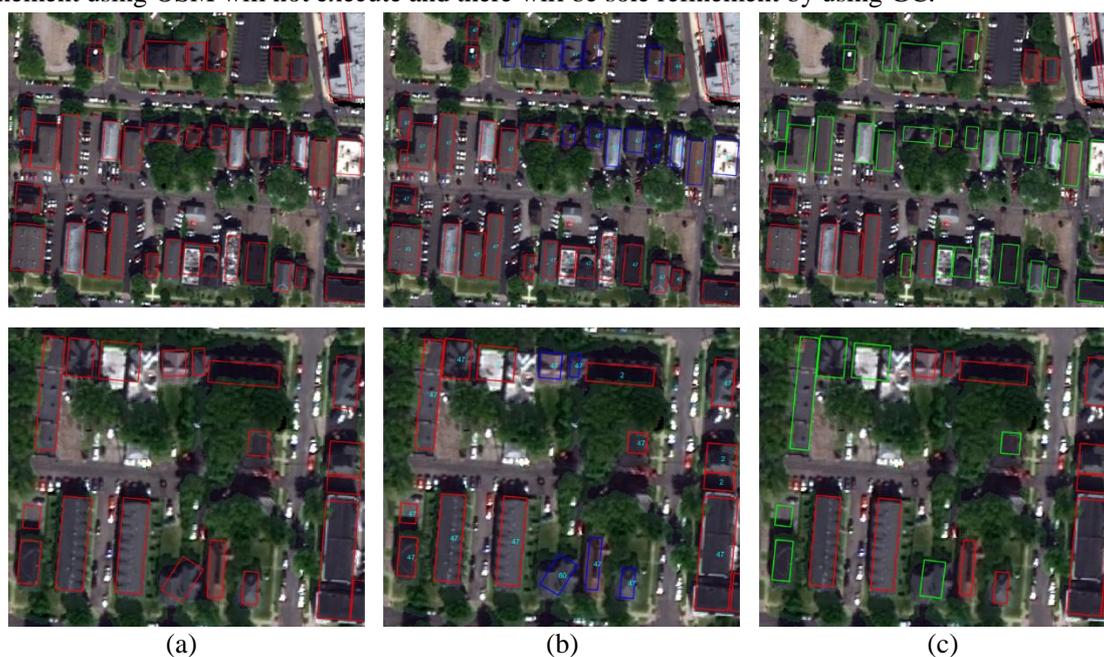

(a)                  (b)                  (c)

**Fig. 6.** Two examples showing building rectangle orientation respectively adjusted by GC (b) and OSM (c).

Blue and Green rectangles refer to those changed during the decomposition. (a) shows polygons before the orientation refinement.

*3.5 3D model fitting*

The extracted building polygons are in rectangle shapes; together with DSM/nDSM, these can be easily fitted by simplex building models. Here we only consider five types of building models namely: flat, gable, hip, pyramid, and mansard, a brief illustration of which is shown in Fig. 7. It can be noted the complexity of the roof topology increases from left to right, along with more parameters to be considered for fitting.

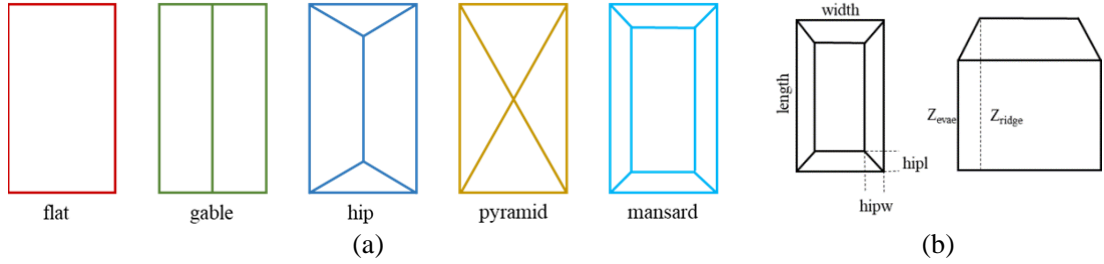

**Fig. 7.** (a) Model roof library with five types and geometrical parameters, color-coded for subsequent demonstration of results. (b). Tunable parameters to define these model types.

Each building model can be described by using several geometrical parameters. we adopt the parameterization based on (Partovi et al., 2019), where the geometrical parameters $\psi$ are defined as follows:

$$\psi \in \Psi; \Psi = \{P, C, S\} \quad (10)$$

where $\Psi$ includes three subsections, $P, C, S$. With $P = \{x_0, y_0, Orientation\}$ are the location parameters of the building model, and $C = \{length, width\}$ are the contour parameters of the building model, and $S = \{Z_{ridge}, Z_{eave}, hipl, hipw\}$ are the shape parameters of the building model, and the geometrically meaning of these parameters are shown in Fig. 7 (b). Each type of building model may be special cases for this type (shown in Fig. 7): for example: for flat building $Z_{ridge} = Z_{eave}$, and for hip building, $hipw = width/2$ etc. Details of different models under this rationale can be found in Table 1, where "width", "length" come from the fitted polygons, and "$\overline{Height}$" is calculated as the mean elevation (from DSM) based on the building polygon (i.e. rectangle).

**Table. 1.** Initial parameters of the building model, parameters with * means these are constants

| Model type | $hipl_{(0)}$ | $hipw_{(0)}$ | $Z_{eave(0)}$ | $Z_{ridge(0)}$ |
|---|---|---|---|---|
| Flat building | 0* | 0* | $\overline{Height} - 0.5$ m | $\overline{Height} - 0.5$ m |
| Gable building | 0* | 1/2 width* | $\overline{Height} - 0.5$ m | $\overline{Height}$ |
| Hip building | 1/4 length | 1/2 width* | $\overline{Height} - 0.5$ m | $\overline{Height}$ |
| Pyramid building | 1/2 length* | 1/2 width* | $\overline{Height} - 0.5$ m | $\overline{Height}$ |
| Mansard building | 1/4 length | 1/4 width | $\overline{Height} - 0.5$ m | $\overline{Height}$ |

The optimization specifically updates the parameter set $S = \{Z_{ridge}, Z_{eave}, hipl, hipw\}$ based on the DSM, where the starting terrain height is computed by taking the local minimum of building height. Given the noisiness of the DSM, we consider directly perform an exhaustive search over the parameters set and choose the model and parameter set with the smallest RMSE. As can be seen from Table 1, the enabled parameters for exhaustive search increase as the model get more complex. We observe that since one building only covers a few hundreds of pixels, even with exhaustive search, the computational time is still reasonable to scale up. Ranges of parameters for searching are shown in Table 2, the value of "searching step size" decided by the resolution of satellite-derived DSM (0.5-meter ground sampling distance (GSD)).

Table. 2. Optimization parameters with their range and step

| Parameter | Parameter range | Searching step size |
|---|---|---|
| $Z_{eave}$ | $(Z_{eave(0)} - 3, Z_{eave(0)} + 3)$ | 0.2 |
| $Z_{ridge}$ | $(Z_{eave} + 0.5, Z_{eave} + 4)$ | 0.2 |
| Hipl | $(hipl_{(0)} - 1/8 \text{ length}, hipl_{(0)} + 1/8 \text{ length})$ | 0.4 |
| Hipw | $(hipw_{(0)} - 1/8 \text{ width}, hipw_{(0)} + 1/8 \text{ width})$ | 0.4 |

*3.6 Model post-refinement*

*3.6.1 Enforcing building type consistency*

Since the type of models are individually fitted and the nosiness of the DSM may result in inconsistent types of building for neighboring and similar building objects (an example shown in Fig. 8(a)). Here we use the same idea to enforce the consistencies of the building types through GC optimization (details in Section 3.4.2), in which label space contains five labels representing the used building types, and the smooth term $B(x_i, x_j)$ keeps the same (Equation (6) ~ (9)) as to enforce label consistency for neighboring buildings with similar color and height, and the data term $R(x_i, \mathcal{L}_i')$ for building type follows a binary representation being a constant if the target label equals to the initial label, otherwise a large number:

$$E(\mathcal{L}') = \sum_{x_i \in P} R(x_i, \mathcal{L}_i') + \lambda \sum_{(x_i, x_j) \in \mathbb{N}} B(x_i, x_j) \delta(\mathcal{L}_i', \mathcal{L}_j') \quad (11)$$

$$R(x_i, \mathcal{L}_i') = 1 - e^{-D(x_i, \mathcal{L}_i')} \quad (12)$$

$$D(x_i, \mathcal{L}_i') = \begin{cases} 0, & Type_{x_i} = Type_{\mathcal{L}_i'} \\ 1, & Type_{x_i} \neq Type_{\mathcal{L}_i'} \end{cases} \quad (13)$$

where $\mathcal{L}' = \{flat, gable, hip, pyramid, mansard\}$ is the label space indicating the building type of the rectangle for type refinement, and $\mathcal{L}_i'$ refers to the optimized label (same as $Type_{\mathcal{L}_i'}$) for building rectangle $x_i$. $Type_{x_i}$ refers to the initial building type of building rectangle and the data term $R(x_i, \mathcal{L}_i')$ tends to remain the same type as initial, the value of which is normalized using an exponential kernel. Data term $R(x_i, \mathcal{L}_i')$ is determined by a binary function $D(x_i, \mathcal{L}_i')$, representing the building type difference between building rectangle $x_i$ and potential label $\mathcal{L}_i'$, and it is set as zero if the building types are the same as the initial type.

To appear in ISPRS Journal of photogrammetry and Remote Sensing (2021)

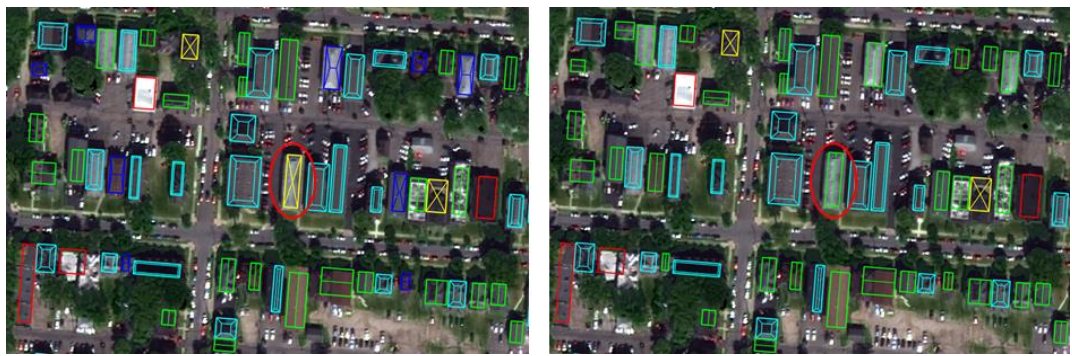

**Fig. 8.** Fitted building types before (left) and after (right) the label consistency enforcement through GC. Different building types are color-coded differently. An example of such a correction is shown in a red circle.

It can be seen from Fig. 8 that the type enforcement through GC has yielded more consistent building types for neighboring buildings. It should be noted this process can be again optional depending on the quality of the DSM.

*3.6.2 Recovering complex 3D buildings through model-level merging*

Large buildings with complex roof structures after decomposition, need to be recovered to 3D topological models. Once individual 2D building rectangles are fitted with one of the five types of models, we consider a model level merging operation to recover potential 3D buildings with complex shapes through line intersection and rule-based merging process, as described below:

To start, we first identify adjacent building rectangles and determine whether or not to perform the merging using criteria as used in Section 3.3 (Equation (3)). The model-level merging is performed at the 2D level and re-fit the model following algorithm from Brédif et al. (2013) (an example of the process is shown in Fig. 9): with two building rectangles (Fig. 9 (a)), it first extends the side of the rectangle and seeks for intersections of these sides (Fig. 9 (b)), followed by the extraction of the enclosed polygon (Fig. 9 (c)) as the base of the more complex shapes. Here we assume the type of the merged polygon may change depending on the original model type of the basic building model. The determination of the merged model types follows heuristics of the building topology, denoted as type conversion matrix shown in Table 3. For example, Pyramid roofs often represent individual buildings and when two of them are identified for merging, it will skip; when one of the two basic building models has the type of pyramid, the type of the resulting merged model will be identified through the converting matrix, as if the pyramid roof has the same type of the other basic building model. Finally, the optimization of the merged roof parameters ($Z_{ridge}, Z_{eave}, hipl, hipw$) follow the method described in Section 3.5 (initial parameters follow the basic model with larger footprint), resulting in the final merged model (Fig. 9 (d)).

**Table. 3.** Roof type decision matrix during model merging

| Roof type | flat | gable | hip | pyramid | mansard |
|---|---|---|---|---|---|
| Flat | flat | gable | hip | flat | mansard |
| Gable | gable | gable | hip | gable | mansard |
| Hip | hip | hip | hip | hip | mansard |
| Pyramid | flat | gable | hip | Not merge | mansard |
| Mansard | mansard | mansard | mansard | mansard | mansard |

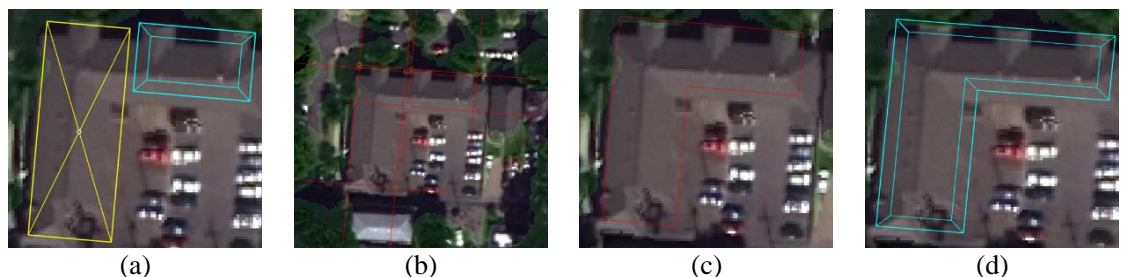

(a) (b) (c) (d)

**Fig. 9.** An example of model-level merging from two basic building models (a); (b) intermediate step for 2D-level merging by extending and intersecting sides of rectangles; (c) merged 2D-level polygon; (d) merged model.

## 4. Experiments

*4.1 Study areas*

The experiments are performed in three cities with different geographical locationsand urban patterns, one located in a typical U.S. city (City of Columbus, Ohio), the other one in South America (Buenos Aires, Argentina), and the last one located in a European Megacity (London, UK). We selected two representative regions for each city (totaling six regions) as shown in Fig. 10 (first row, overlaid with ground-truth labels)**:** The Columbus-area-1 shows a typical residential/commercial region with relatively dense buildings in similar sizes and Columbus-area-2 a typical industrial region containing buildings with varying sizes; Buenos-Aires-area-1 presents a very challenging scene that contains both large-sized building and small & dense buildings; Buenos-Aires-area-2 presents a region with relatively sparse and single houses, with disturbances such as swimming pools and regular playgrounds; The London-area-1 shows a typical block region with compact buildings; London-area-2 shows a region with numerous complex buildings. Each regions covers approximately 0.5 to 2.25 km² area with Orthophoto and DSM at a 0.5 m GSD (ground sampling distance). The orthophoto and DSMs are produced using a multi-view stereo approach (Qin, 2019; Qin, 2017; Qin, 2016) based on five worldview-3 stereo pairs for Buenos Aires dataset and 12 World-view stereo pairs for the Columbus dataset. The accuracy of the DSMs was analyzed systematically in the work of (Qin, 2019), which was reported to have achieved sub-meter vertical accuracy in terms of RMSE (root-mean-squared error). The orthophotos are pan-sharpened using an 8-band multispectral image, while for simplicity, converted to 3-band RGB for building detection and processing. The evaluation of the geometry and detection are evaluated separately using both a 2D Intersection over Union (IOU2) and 3D Intersection over Union (IOU3) based on manually created reference data for building footprint and LiDAR-based DSM for 3D geometry (as shown in Fig. 10). The IOU2 and IOU3 are defined following (Kunwar et al., 2020) as follows:

$$IOU2 = \frac{TP}{TP+FP+FN} \quad (14)$$

$$IOU3 = \frac{TP_{3D}}{TP_{3D}+FP+FN} \quad (15)$$

where $TP$ is the number of true positives pixels that is determined as extracted and manually labeled building footprint simultaneously, $FP$ is the number of false positives and $FN$ is the number of false negatives. $TP_{3D}$ is $TP$ pixels whose 3D vertical difference from the ground-truth LiDAR is within 2 m.

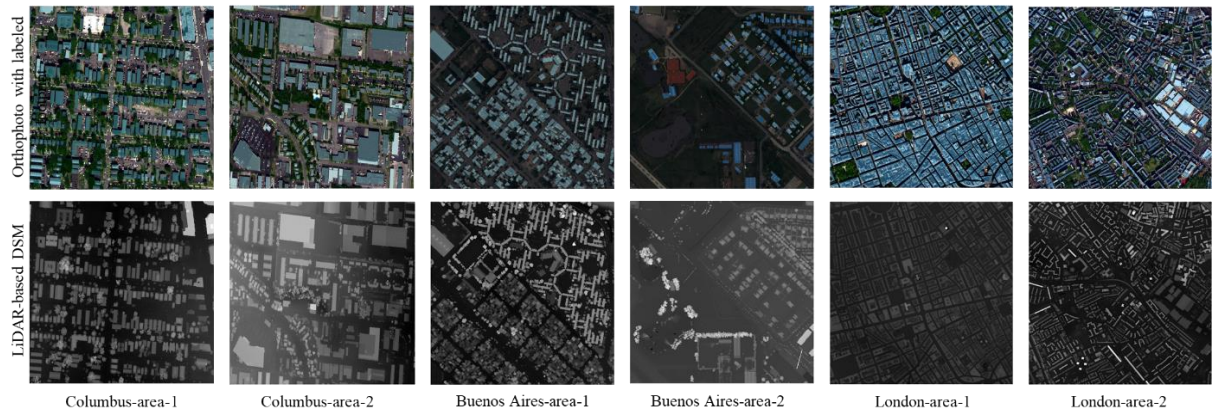

**Fig. 10.** First row: Orthophotos of the study areas, overlaid with manually drawn masks; second row: corresponding ground-truth LiDAR data.

*4.2 Experimental results*

The orthophotos study areas are shown in Fig. 10, first row, overlaid with manually drawn building masks results of the building 2D polygon and rectangle extraction (methods described in Section 3.1-3.4) and building 3D model fitting (methods described in Section 3.5-3.6) are shown in Fig. 11, which specifically includes building mask detection, initial building polygon detection, decomposition & merging, and orientation refinement with GC and OSM. We show the model fitting results of the six experimental regions in the first row of Fig. 11, by projecting the wireframes of the model on the orthophoto; in the second row of Fig. 11, we demonstrate intermediate results of the first region (i.e. Columbus-area-1) including initial building polygon extraction, building rectangle decomposition and refinement, as well as the final fitted more, with the third row of Fig. 11 highlighting some of the results. It can be seen that the proposed method has detected most of the buildings and correctly outlined the building boundaries. In addition, most of the buildings initially detected to be connected are successfully decomposed as individual rectangles. Minor errors are observed for small and complex buildings where decompositions may fail: an example is shown in the third row (circled in red), which has erroneously separated buildings into a thin rectangle thus being reconstructed incorrectly.

Fig. 11 also shows the final result of building modeling for the six regions with roof type, and classes other than building are set as terrain DSM. Fig. 12 display the LoD-2 reconstructed buildings in six experimental areas. The result indicates that the proposed approach performs well on community building in Columbus-area-1, Columbus-area-2, and Buenos-Aires-area-2, while in extraordinarily complex blocks, like Buenos-Aires-area-1 and London-area-1, some buildings are not reconstructed so accurate, since building mask from Section 3.1 does not detect perfect segments.

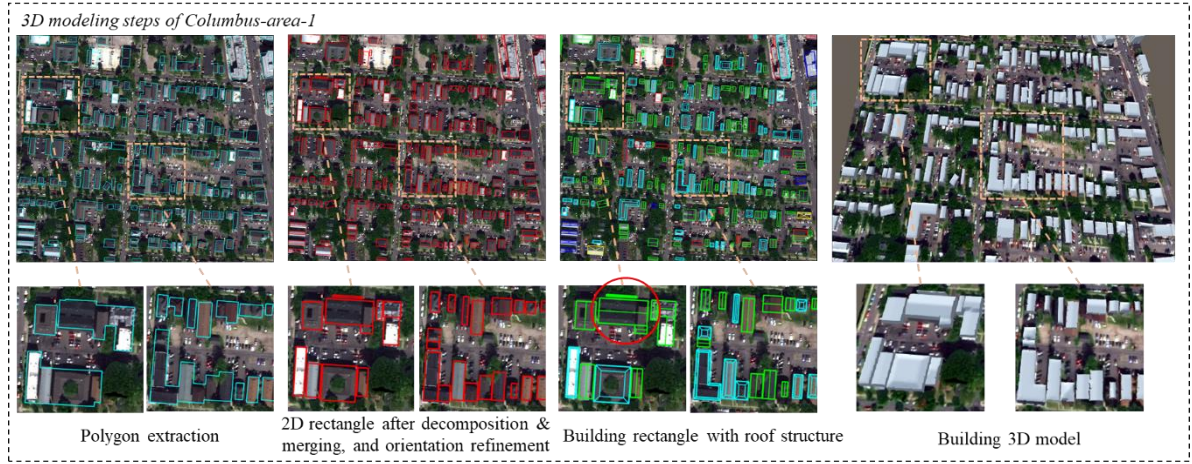

**Fig. 11.** LoD-2 model reconstruction steps for Columbus-area-1. First row: Intermediate results of the "Columbus-area-1" region including initial building polygon extraction, polygon decomposition and refinement, model fitting and the last figure of this row shows the 3D visualization. The second row enlarges part of each figure of the second row for visualization. The building circled in red shows an example of erroneous reconstruction.

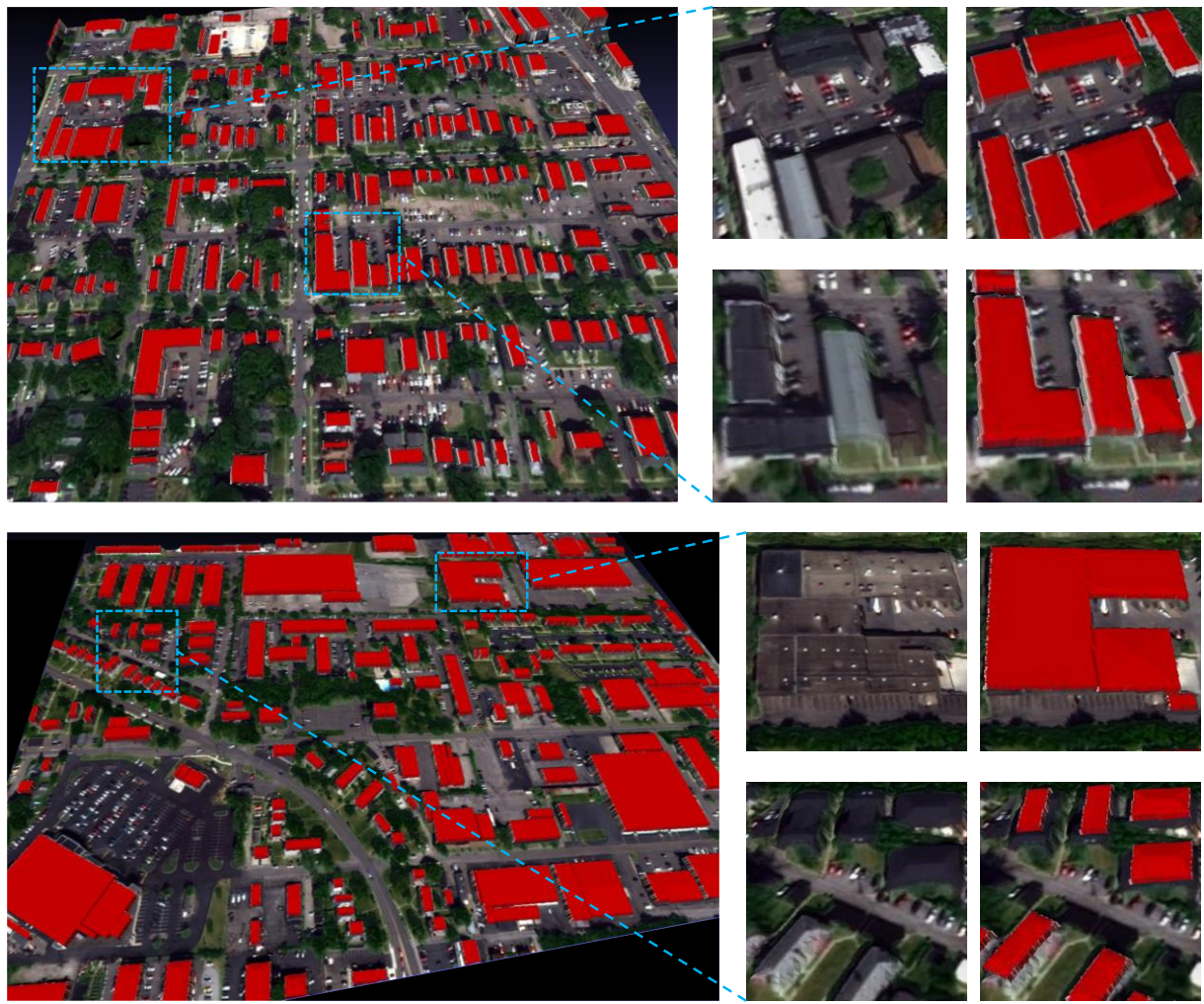

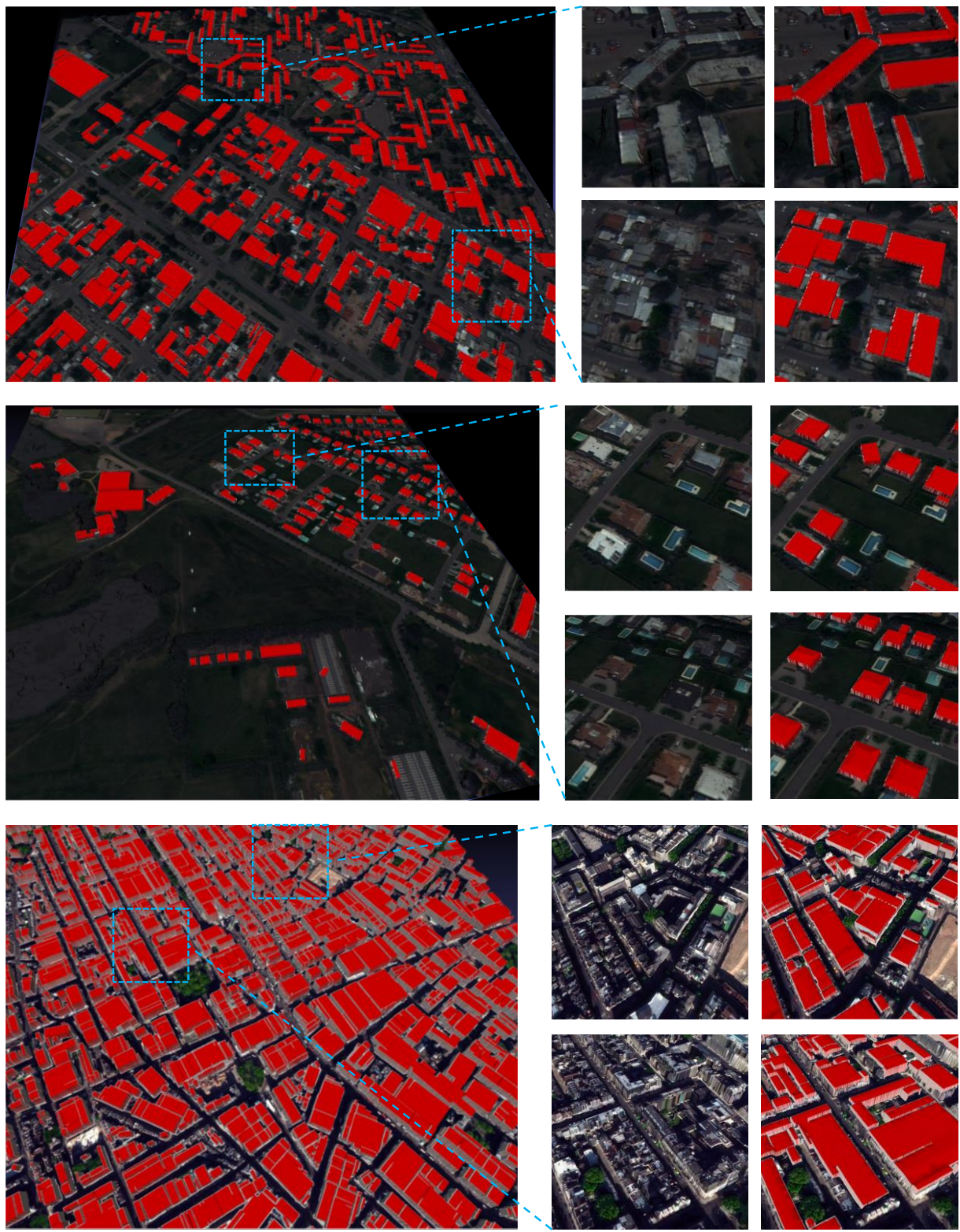

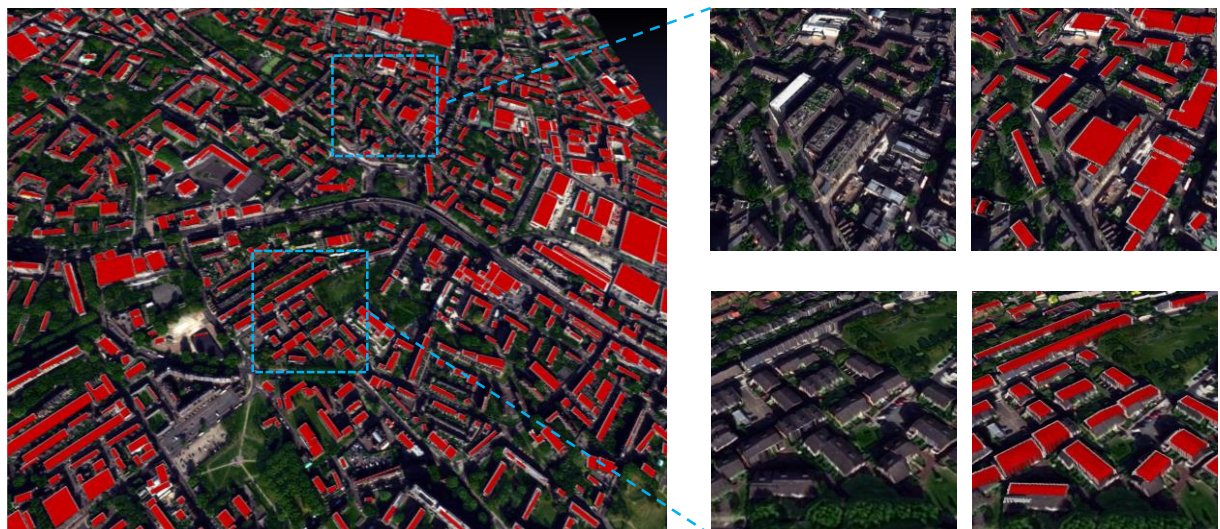

**Fig. 12.** LoD-2 model reconstruction results of the six experimental regions. Left column: 3D model with roof covered by red mask and façade covered by white; medium column: DSM with texture in a close view, right column: 3D model with roof covered by red mask in a close view. From top to bottom: Columbus-area-1, Columbus-area-2, Buenos-Aires-area-1, Buenos-Aires-area-2, London-area-1, and London-area-2.

*4.3 Accuracy evaluation of experiment areas*

The accuracy of the resulting models is evaluated using the IOU2 and IOU3 metrics, respectively to assess the 2D building footprint extraction accuracy and the 3D fitting accuracy. The accuracy of 'DSM' represents the raw measurements to be compared with the ground truth. We have ablated the results with and without the GC and OSM refinements, statistics against the ground truth are shown in Table 4, where "OSM" or "GC" indicates that the orientation is refined using OSM or GC alone, and "OSM+GC" indicates the orientation being refined by OSM first and followed by GC. Notable examples are shown in Fig. 13, the first two rows of which indicated that the OSM might be able to correct orientations of buildings, while do not influence cases if orientations of the buildings are already well-estimated (third row of Fig. 13). In both cases, OSM and GC refinement had a positive impact on the resulting metrics in three out of the six areas, and it is possible that incorrect OSM might adversely impact the final IOU. It should be noted that although the statistics show only a marginal improvement of the GC and OSM refinement (largely due to that these adjustments are small and the building segments are rather accurate), it visually shows a much better consistency in occasions where the buildings are misoriented (e.g. Fig. 6). The fitted DSM has a larger error distribution than the original DSM, since often fitting will result in reduced accuracy given that the regularized shape may introduce errors, as for example, a raw measurement curved surface may be approximated by multiple pieces of a planar surface, and the same applies to the model fittings.

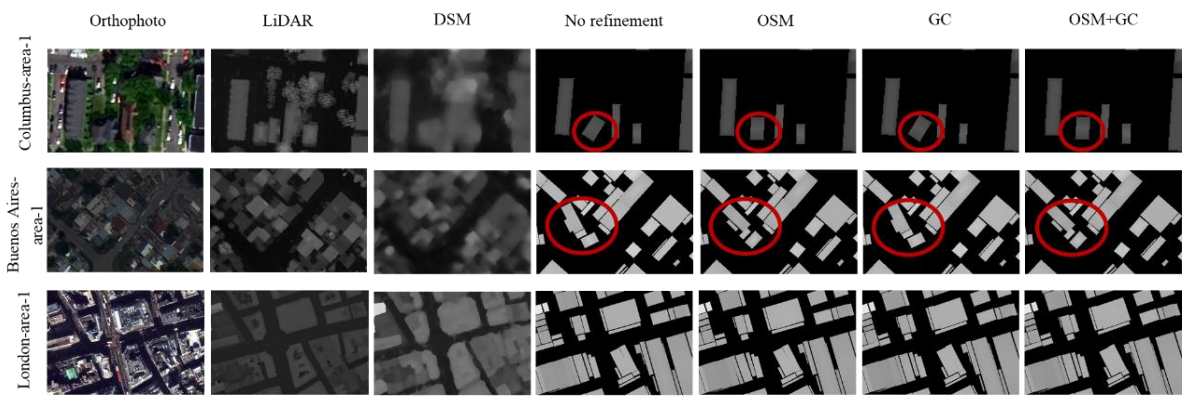

**Fig. 13.** Building model height result with different orientation refinement options. The red circled regions highlight notable differences among different strategies.

**Table. 4.** Accuracy evaluation of the resulting building models

| Region | Accuracy | DSM | No refinement | OSM | GC | OSM+GC |
|---|---|---|---|---|---|---|
| Columbus area 1 | IOU2 | 0.6929 | 0.6348 | *0.6359* | 0.6349 | **0.6362** |
| | IOU3 | 0.6632 | 0.5780 | *0.5794* | 0.5782 | **0.5796** |
| Columbus area 2 | IOU2 | 0.8360 | 0.8076 | *0.8084* | 0.8076 | **0.8085** |
| | IOU3 | 0.8287 | 0.7950 | *0.7961* | 0.7951 | **0.7962** |
| Buenos Aires area 1 | IOU2 | 0.6709 | 0.6378 | 0.6377 | *0.6380* | **0.6381** |
| | IOU3 | 0.6551 | 0.5892 | 0.5888 | *0.5893* | **0.5894** |
| Buenos Aires area 2 | IOU2 | 0.3539 | *0.5301* | 0.5298 | **0.5302** | *0.5301* |
| | IOU3 | 0.3281 | **0.4829** | 0.4825 | *0.4828* | 0.4827 |
| London area 1 | IOU2 | 0.6830 | 0.5797 | **0.5799** | 0.5796 | *0.5798* |
| | IOU3 | 0.6176 | 0.4711 | **0.4714** | 0.4707 | *0.4712* |
| London area 2 | IOU2 | 0.6004 | **0.5053** | *0.5052* | 0.5048 | *0.5052* |
| | IOU3 | 0.5626 | **0.4180** | **0.4180** | 0.4160 | 0.4163 |

*4.4 Comparative study*

We compare our results with results generated by state-of-the-art methods, and given the nature of the LoD-2 model reconstruction method being component-rich, trivial and often ad-hoc, we re-implement and compare the key components of some of the existing methods on their algorithms for building polygon extraction and decomposition, these methods being: the method in (Partovi et al., 2019), (Wei et al., 2020), (Arefi & Reinartz, 2013), and that in (Li et al., 2019). The two key components we are comparing against existing methods are: 1) building polygon extraction (process described in Section 3.2), and 2) building polygon decomposition (process described in Section 3.3). The building polygon extraction methods include 1) a generalization of line segments-based building outline extraction method (Partovi et al., 2019), which generates building mask by applying SVM classification to gradient feature from PAN image and extract primitive boundary points by using SIFT algorithm (Lowe, D. G, 2004), and then fit boundary line segments from points and regularized them by finding building's orientation; 2) a toward automatic building footprint delineation by using CNN and regularization method (Wei et al., 2020), which firstly segments buildings via FCN with multiple scale aggregation of feature pyramids from convolutional layers, and next regularizes polygon by adapting a coarse and fine polygon adjustment; 3) a minimum bounding rectangle (MBR) based method (Arefi & Reinartz, 2013), which approximates the remaining polygon by calculating bounding rectangle of building segment and the difference mask between bounding rectangle and building mask, with MBR-based algorithm for rectilinear building and RANSAC-based approximation algorithm for non-rectilinear building. The rectangle decomposition methods include 1) a parallel line-based building decomposition method (Partovi et al., 2019), which moves line segments until it meets the buffer of another parallel line segment and then generates rectangles using these two parallel line segments; 2) a primitive-based 3d building modeling method (Li et al., 2019), which cascades a building into a set of parts via mask R-CNN, and uses a greedy approach to select and move instance with largest IOU to decompose the building into a set of shapes. To ensure that the performance of building polygon extraction methods and rectangle decomposition method are evaluated individually, the comparison is designed as that polygon extraction methods share the same building mask input from Section 3.1, and rectangle decomposition methods share the same building polygon from Section 3.2 and IOU2 is used as the metric for evaluation. The building polygon extraction from (Partovi et al., 2019) was realized using the SIFT algorithm to detect key points and those points are selected as boundary points only around the building mask from Section 3.1. For (Wei et al., 2020) and (Arefi & Reinartz, 2013), primitive building segment is replaced by using building mask from Section 3.1.

  Fig. 14 shows the sample result of all methods for building polygon extraction, Fig. 15 shows the sample result of all methods for building polygon decomposition to building rectangles, and Table 5 gives the statistics. It can be seen that our method in both of the tasks outperforms all the existing methods, especially in Fig. 14-15**,** our methods due to the nature of deeply integrating decision criteria on the image color similarity and DSM continuity through a grid-based approach, the detection and decomposition algorithms are able to identify regularized polylines, and identify separating boundaries for building rectangles. Compared to all other existing approaches, either miss detections or incorrectly locate or identify building rectangles. Examples showing the limitation of our method can be found in Fig. 15**,** the last row, in which our method has failed to separate a connected building and created a minor artifact due to shadows, whereas other methods in this example are found to be worse in decomposition. Table 5 shows that among the six regions we have experimented on, the IOU2 of our results achieves the highest for both tasks, except for Buenos-Aires-area 2 and London-area 2, which is marginally lower.

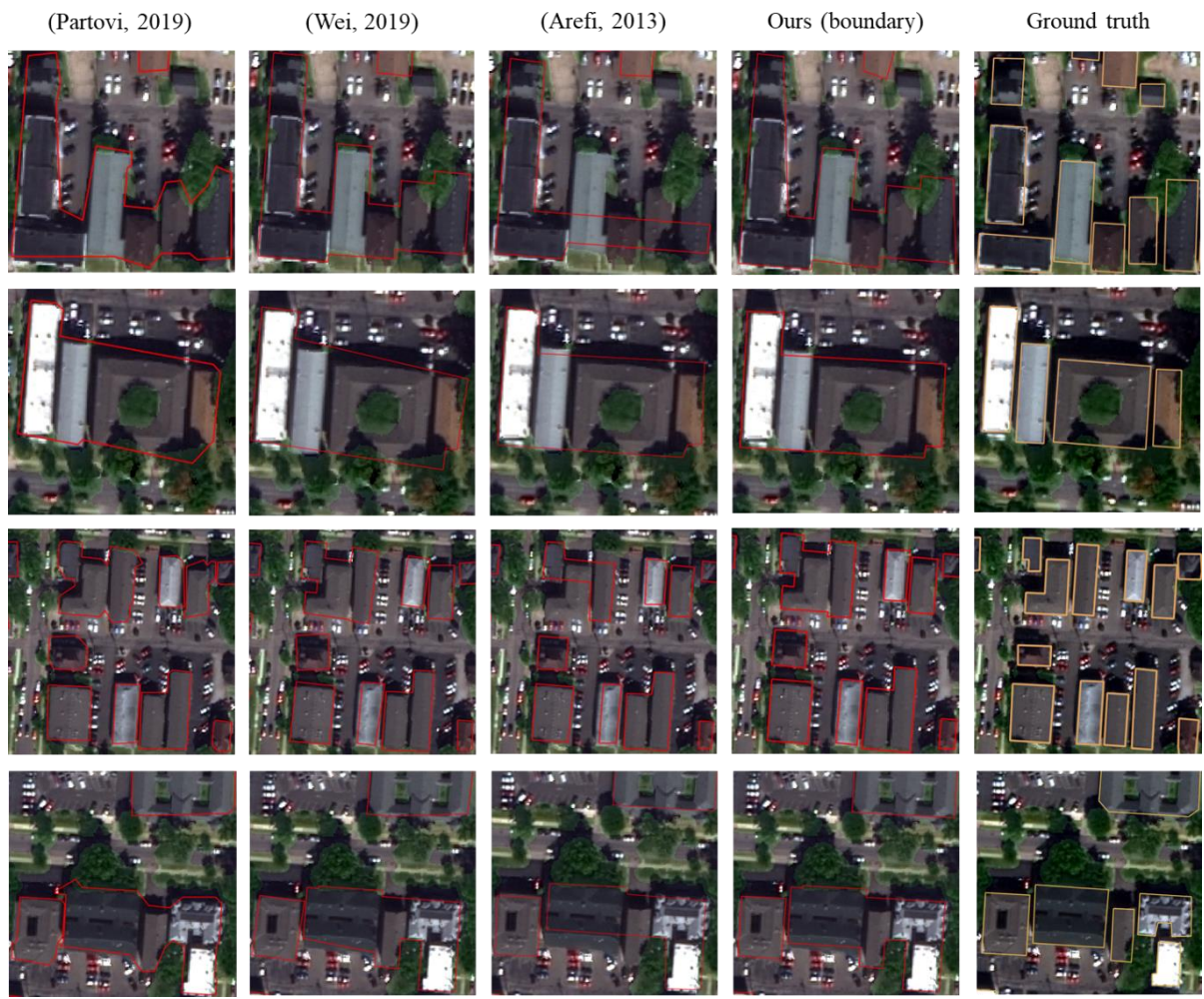

**Fig. 14.** Comparative results on building polygon extraction with other state-of-the-art methods.

**Table. 5.** IOU2 values of comparing results in building footprint extraction and decomposition (Bold values indicate the best performing approach and italic the second best)

|  | Method | Columbus area1 | Columbus area2 | Buenos Aires area 1 | Buenos Aires area 2 | London area 1 | London area 2 |
|---|---|---|---|---|---|---|---|
| Building Polygon Extraction | (Partovi, 2019) | *0.6827* | 0.8157 | 0.6456 | **0.6109** | *0.6623* | 0.5744 |
|  | (Wei, 2020) | 0.6811 | *0.8179* | *0.6543* | 0.6093 | 0.6601 | **0.5765** |
|  | (Arefi, 2013) | 0.6433 | 0.7761 | 0.5381 | 0.5778 | 0.4545 | 0.4639 |
|  | Ours | **0.6831** | **0.8195** | **0.6549** | *0.6099* | **0.6624** | *0.5749* |
| Rectangle Decomposition | (Partovi, 2019) | *0.6579* | *0.7679* | 0.5976 | 0.5532 | 0.5745 | 0.4848 |
|  | (Li, 2019) | 0.5768 | 0.7618 | *0.5980* | *0.5533* | *0.5781* | *0.5110* |
|  | Ours | **0.6587** | **0.8133** | **0.6375** | **0.5969** | **0.5796** | **0.5136** |

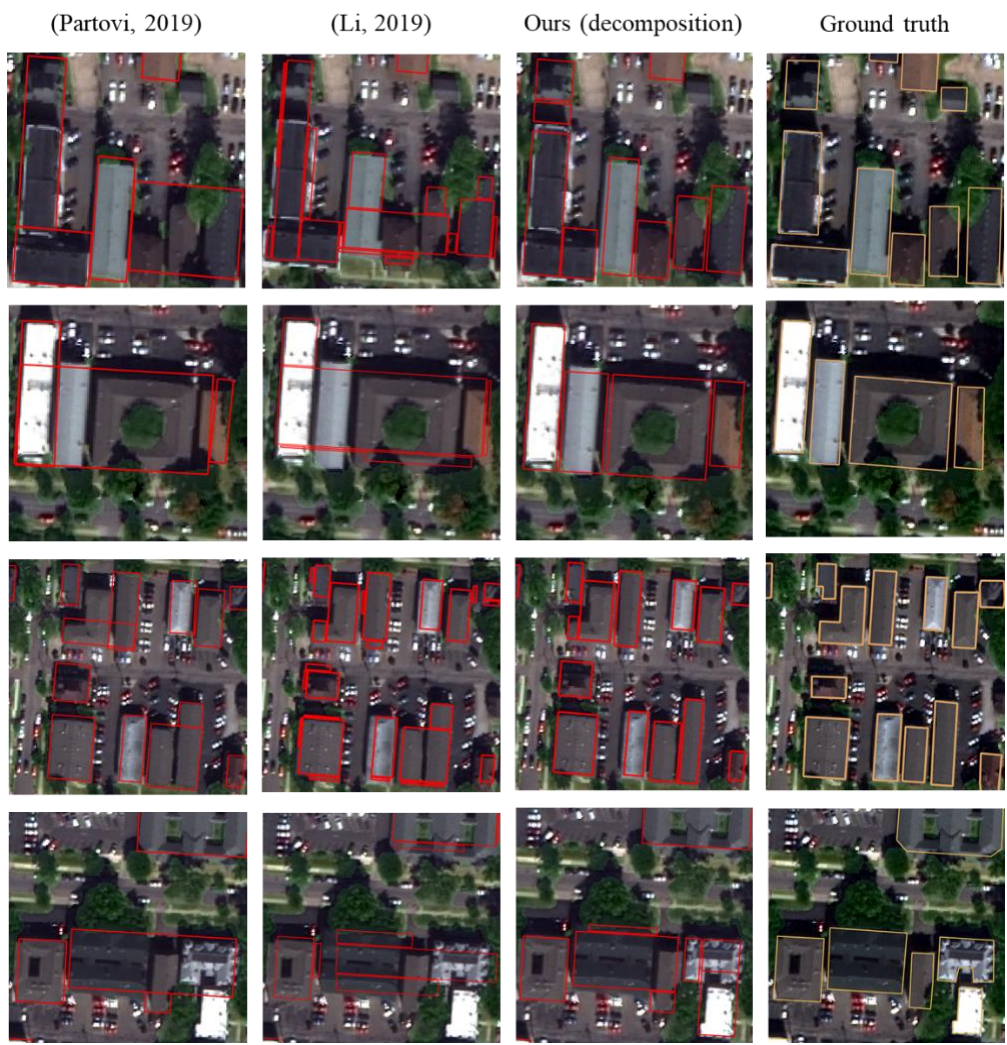

**Fig. 15.** Comparative results on building polygon decomposition with other state-of-the-art methods.

*4.5 Parameter analysis*

With several tunable parameters in the proposed approach, it is worth discussing the sensitivity of the parameters in Table 6. All five thresholds listed in Table. 6 contribute to the 2D building shape in the approach, and are assigned based on the empirical studies in the experiment. Two areas in Columbus are taken to analyze the impact of each threshold. A pair of initial thresholds are separately assigned to those two experimental areas, with $\boldsymbol{T_1} = \{T_w, T_l, T_d, T_{h1}, T_{h2}\} = \{0.16, 120, 10, 1, 0.2\}$ for Columbus-area-1 and $\boldsymbol{T_2} = \{T_w, T_l, T_d, T_{h1}, T_{h2}\} = \{0.2, 120, 10, 1, 0.2\}$ for Columbus-area-2. IOU2 of building rectangles after building polygon decomposition is calculated to represent the performance of different pairs of thresholds. The initial IOU2 of Columbus-area-1 with threshold $\boldsymbol{T_1}$ equals 0.6584, and 0.8131 for Columbus-area-2 with threshold $\boldsymbol{T_2}$. Fig. 16 shows the relationship and difference between IOU2 with individually changed parameters $\boldsymbol{T'}$ and IOU2 with initial parameters $\boldsymbol{T_1}$ and $\boldsymbol{T_2}$, equals to $\boldsymbol{T'} - \boldsymbol{T_i}$, with $\boldsymbol{T'}$ means the pair of threshold that solely change one threshold based on $\boldsymbol{T_i}$, and $i = \{1, 2\}$. Fig. 16(a) shows the decreasing trend of IOU2 when the weight threshold $T_w$ close to 0.1 in Columbus-area-1; Fig.16(b) to (e) show that the tuning thresholds lead to an influence lower than 0.005 of IOU2, which represents there are minors influences once those thresholds are assigned as different values. Therefore, the sensitivity of several major thresholds is reliable for the proposed approach.

**Table. 6.** List of tunable parameters of the proposed approach

| Parameter | Section | Description |
|---|---|---|
| Weight threshold $T_w$ | Section 3.1.3 | A threshold of decision weight of a bounding box, to determine whether to use building segment from Mask R-CNN. |
| Length threshold $T_l$ (pixel) | Section 3.2 | A threshold of summed up length for determining building main orientations. |
| Color difference threshold $T_d$ (RGB) | Section 3.3 | A threshold of mean color differences ($|\overline{C_1} - \overline{C_2}|$) of the two rectangles to decide whether to merge two nearby rectangles in building decomposition. |
| Mean height difference threshold $T_{h1}$ (m) | Section 3.3 | A threshold of mean height difference ($|\overline{H_1} - \overline{H_2}|$) between two nearby rectangles to decide whether to merge two nearby rectangles in building decomposition. |
| Gap threshold $T_{h2}$ (m) | Section 3.3 | A threshold of dramatic height changes in a buffered region that cover the common edge between two nearby rectangles between two nearby rectangles. |

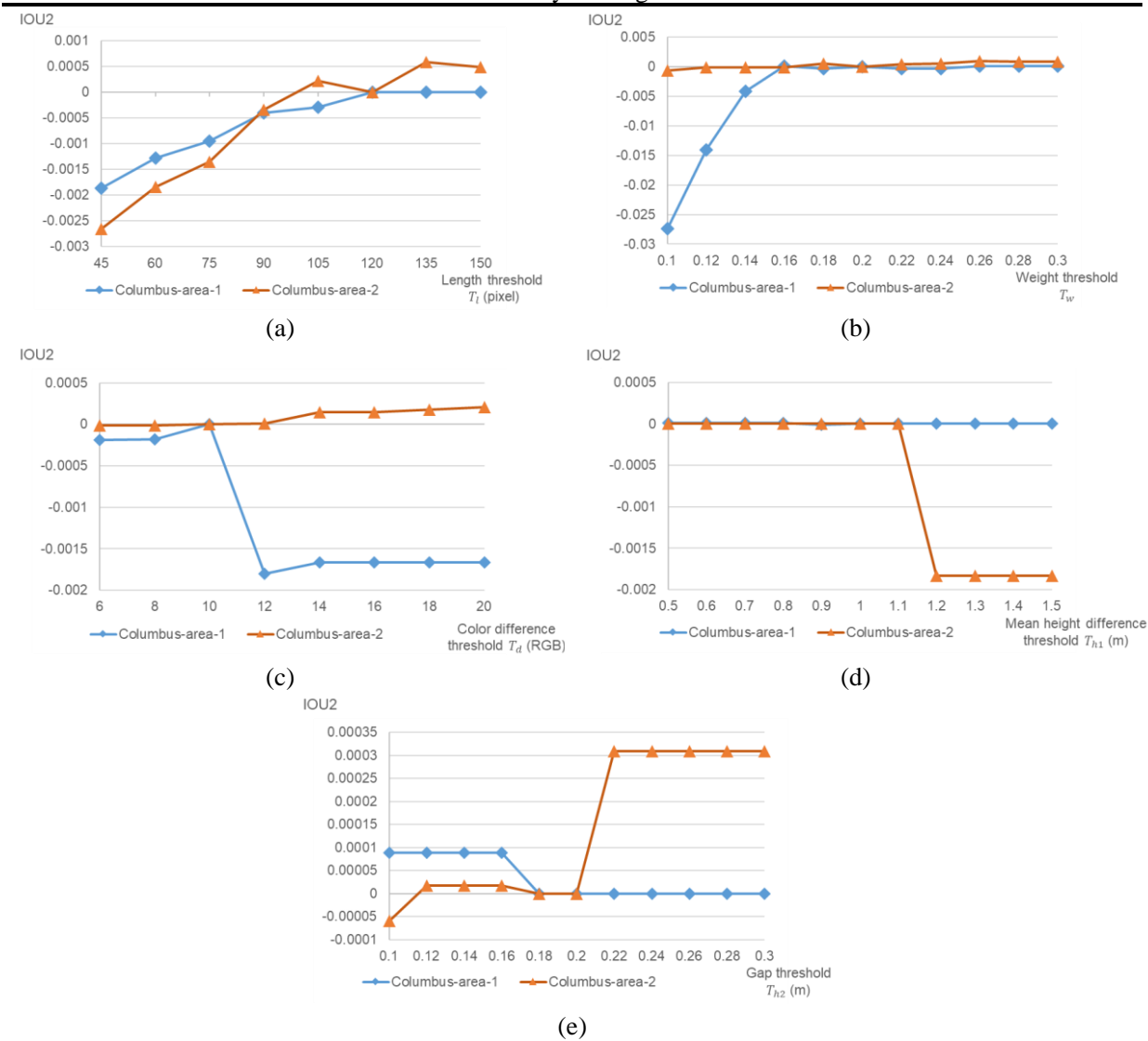

(a)

(b)

(c)

(d)

(e)

**Fig. 16.** The IoU2 of decomposed building rectangles differenced between initial setting and changed thresholds. Columbus-area-1 initial $IOU2 = 0.6584$, Columbus-area-2 initial $IOU2 = 0.8131$. (a) the relationship between weight threshold $T_w$ and IOU2; (b) the relationship between length threshold $T_l$ and IOU2; (c) the relationship between color difference threshold $T_d$ and IOU2; (d) the relationship between mean height difference threshold $T_{h1}$ and IOU2; (e) the relationship between gap threshold $T_{h2}$ and IOU2;

## 5 Conclusion

In this paper, we propose an
 LoD-2 model reconstruction approach performed on DSM and orthophoto derived from very-high-resolution multi-view satellite stereo images (0.5 meter GSD). The proposed method follows a typical model-driven paradigm that follows a series of steps including: instance-level building segment detection, initial 2D building polygon extraction, polygon decomposition and refinement, basic model fitting and merging, in which we address a few technical caveats over existing approaches: 1) we have deeply integrated the use of color and DSM information throughout the process to decide the polygonal extraction and decomposition to be context-aware (i.e., decision following orthophoto and DSM edges); 2) a grid-based decomposition approach to allow only horizontal and vertical scanning lines for computing gradient for regularized decompositions (parallelism and orthogonality). Six regions from two cities presenting various urban patterns are used for experiments and both IOU2 and IOU3 (for 2D and 3D evaluation) are evaluated, our approaches have achieved an IOU2 ranging from 47.12% to 80.85%, and an IOU3 ranging from 41.46% to 79.62%. Our comparative studies against a few state-of-the-art results suggested that our method achieves the best performance metrics in IOU measures and yields favorably visual results. Furthermore, our parameter analysis indicates the robustness of threshold tuning for the proposed approach.

Given that our method assumes only a few model types rooted in rectangle shapes, the limitation is that the proposed approach may not perform for other types of buildings such as those with dome roofs and may to over-decompose complex-shaped buildings. It should be noted the proposed method involves a series of basic algorithms that may involve resolution-dependent parameters, and default values are set based on 0.5 meter resolution data and can be appropriately scaled when necessarily processing data with higher resolution, while the authors suggest when processing with higher resolution data that are potentially sourced from airborne platforms, bottom-up approaches or processing components can be potentially considered to yield favorable results. The proposed approach developed in this paper, is specifically designed for satellite-based data that rich the existing upper limit of resolution (0.3-0.5 GSD) to accommodate the data uncertainty and resolution at scale. In the region with numerous compact blocks, the proposed approach capability is limited to reconstruct the roof structure of those blocks.

In our future work, a direct prediction of model type and parameters will be attempted, and other building segmentation methods will be introduced for building mask improvement, and types of models will be increased rooted not only on rectangle shapes but also circular and complexly parameterized shapes, followed by continued investigation on approaches to favorably offer reasonable decomposition of overcomplex building and post-merging. In addition, as future works it is worth establishing benchmark datasets with varying sources, where LiDAR data are available to construct LoD-2 ground truth data, which can evaluate image-based building model reconstruction approaches.

## 6. Acknowledgements

This work is supported in part by the Office of Naval Research (Award No. N000141712928). Part of the datasets used and involved in this research were released by IARPA, John Hopkins University Applied Research Lab, IEEE GRSS committee, and SpaceNet. The satellite imagery in the MVS benchmark data set was provided courtesy of DigitalGlobe.